# Myths and Legends of the Baldwin Effect

Peter Turney
Institute for Information Technology
National Research Council Canada
Ottawa, Ontario, Canada, K1A 0R6
peter@ai.iit.nrc.ca

## Abstract

This position paper argues that the Baldwin effect is widely misunderstood by the evolutionary computation community. The misunderstandings appear to fall into two general categories. Firstly, it is commonly believed that the Baldwin effect is concerned with the synergy that results when there is an evolving population of learning individuals. This is only half of the story. The full story is more complicated and more interesting. The Baldwin effect is concerned with the costs and benefits of lifetime learning by individuals in an evolving population. Several researchers have focussed exclusively on the benefits, but there is much to be gained from attention to the costs. This paper explains the two sides of the story and enumerates ten of the costs and benefits of lifetime learning by individuals in an evolving population. Secondly, there is a cluster of misunderstandings about the relationship between the Baldwin effect and Lamarckian inheritance of acquired characteristics. The Baldwin effect *is not* Lamarckian. A Lamarckian algorithm *is not* better for most evolutionary computing problems than a Baldwinian algorithm. Finally, Lamarckian inheritance *is not* a better model of memetic (cultural) evolution than the Baldwin effect.

## 1 Introduction

Since Hinton and Nowlan's (1987) classic paper, several researchers have observed a synergetic effect in evolutionary computation when there is an evolving population of learning individuals (Ackley and Littman, 1991; Belew, 1989; Belew *et al*., 1991; French and Messinger, 1994; Hart, 1994; Hart and Belew, 1996; Hightower *et al*., 1996; Whitley and Gruau, 1993; Whitley *et al*., 1994). This synergetic effect is usually called *the Baldwin effect*. This has produced the misleading impression that there is nothing more to the Baldwin effect than synergy. A myth or legend has arisen that the Baldwin effect is simply a special instance of synergy. One of the goals of this paper is to dispel this myth.

Roughly speaking (we will be more precise later), the Baldwin effect has two aspects. First, lifetime learning in individuals can, in some situations, accelerate evolution. Second, learning is expensive. Therefore, in relatively stable environments, there is a selective pressure for the evolution of instinctive behaviors. Recent research in evolutionary computation has focussed almost exclusively on the first aspect of the Baldwin effect. This paper is an attempt to encourage a more balanced view. Learning has benefits (the first aspect of the Baldwin effect) but it also has costs (the second aspect). The Baldwin effect is concerned with the costs and benefits of lifetime learning in an evolving population.

The second goal of this paper is to clarify the relationship between the Baldwin effect and Lamarckian inheritance of acquired characteristics. The Baldwin effect resembles Lamarckism in some ways: behaviors that are learned in one generation become instinctive in a later generation. However, the Baldwin effect is purely Darwinian. Unlike Lamarckism, acquired characteristics (behaviors that an individual acquires by lifetime learning) are not directly inherited. What is inherited is the ability to acquire the characteristics (the ability to learn).

We begin in Section 2 with a brief discussion of terminology. Section 3 attempts to accurately describe the Baldwin effect. We then discuss the costs and benefits of lifetime learning in an evolving population in Section 4. The relationship between Lamarckism and the Baldwin effect is examined in Section 5. We conclude in Section 6.

## 2  Terminology

Before we continue, we should define a few terms. The *genotype* is the genetic constitution of an individual. In a living organism, this is typically the organism's DNA. In evolutionary computation, it is typically a string of bits. The *phenotype* is the set of observable characteristics of an organism, as determined by the organism's genotype and environment. Roughly speaking, the genotype is the DNA and the phenotype is the body. The distinction between genotype and phenotype is clear in biological evolution, but the distinction does not exist in many of the simpler examples of evolutionary computation.

*Lifetime learning* is learning during the lifetime of an individual. The evolution of a species may also be viewed as a form of learning, but it is learning at the level of populations, while lifetime learning is learning at the level of the individual. *Phenotypic plasticity* is the ability of an organism to adapt to its environment. The most obvious form of phenotypic plasticity is lifetime learning. However, there are many other forms of phenotypic plasticity, such as our ability to tan in sunny environments or our ability to form a callus when our skin is repeatedly abraded. In its most general sense, the Baldwin effect deals with the impact of phenotypic plasticity on evolution. The impact of lifetime learning on evolution is only one example of the Baldwin effect.

## 3  The Baldwin Effect

The "Baldwin effect" is a misnomer, because the Baldwin effect was discovered independently by Baldwin (1896), Morgan (1896), and Osborn (1896), and also because it is not a single effect. It is rather a cluster of effects, or perhaps a cluster of observations.

**1. Benefits of phenotypic plasticity:** Briefly, phenotypic plasticity smooths the fitness landscape, which can facilitate evolution. In more detail, phenotypic plasticity enables the organism to explore neighboring regions of phenotype space. The fitness of an organism is then determined (approximately) by the maximum fitness in its local region of phenotype space. If the genotype and the phenotype are correlated, so that a small change in one usually corresponds to a small change in the other, then the fitness of the genotype of a plastic individual is given (approximately) by the maximum fitness in the local region in genotype space. Therefore, plasticity has the effect of smoothing the fitness landscape. This makes it easier for evolution to climb to peaks in the landscape.

**2. Benefits of phenotypic rigidity:** Phenotypic rigidity can be advantageous in many situations. Therefore organisms may slowly evolve rigid mechanisms that replace or augment their plastic mechanisms. (The term "mechanism" is intended to include both behaviors and physical structures.) For example, learning requires experimentation, which can be dangerous. There can be advantages to instinctively avoiding snakes, instead of learning this behavior by trial-and-error. Similarly, it takes time to build a callus. There can be advantages to being born with thickened skin on the palms and soles.

**3. Plasticity of behavior:** Behaviors tend to be more plastic than physical structures. To learn a new behavior, an organism must make changes to its nervous system. The nervous system tends to be more flexible and adaptable than other structures in the body. For example, tanning is one way to adapt to the sun. Learning provides us with many other ways to adapt: we may seek shade, wear clothes, or use sun screen.

**4. Plasticity of learning:** Learned behaviors tend to be more plastic than instinctive behaviors. That is, the learned-instinctive continuum is an instance of the plastic-rigid continuum. Instincts are part of how we adapt to our environment, but learning is more flexible than instinct; learning allows adaptation to a wider range of environments.

Most of the work in the artificial life and genetic algorithm communities has focused on the benefits of phenotypic plasticity and the plasticity of learning (observations 1 and 4) (Ackley and Littman, 1991; Belew, 1989; Belew *et al.*, 1991; French and Messinger, 1994; Hart, 1994; Hart and Belew, 1996; Hightower *et al.*, 1996; Whitley and Gruau, 1993; Whitley *et al.*, 1994; Balakrishnan and Honavar, 1995; Belew and Mitchell, 1996). Together, these observations imply that learning can facilitate evolution.

Some recent work in ALife and GA has combined analysis of the benefits of phenotypic plasticity, phenotypic rigidity, and the plasticity of learning (observations 1, 2, and 4) (Anderson, 1995a, 1995b; Cecconi, 1995; Hinton and Nowlan, 1987; Behera and Nanjundiah, 1995). These observations imply that learning can facilitate evolution, but learned behaviors may eventually be replaced by instinctive behaviors.

Biologists have focused on analysis of the benefits of phenotypic plasticity, phenotypic rigidity, and the plasticity of behavior (observations 1, 2, and 3) (Wcislo, 1989; Maynard Smith, 1987; Waddington, 1942; Scheiner, 1993; Simpson, 1953; West-Eberhard, 1989; Gottlieb, 1992). These observations imply that behaviors can facilitate the evolution of physical structures.

The biological work (observations 1, 2, 3) makes an interesting contrast to the GA/ALife work (observations 1, 2, 4). Many computer simulations have modelled the distinction between learning and instinct, but (as far as we know) no simulations have modelled the distinction between physical structure and behavior. It is not clear that this distinction has any meaning in a computer simulation. It appears to require evolving physical robots.

Our focus in this paper will be on learning versus instinct, since our primary interest is in evolutionary computation. Most of what we say here about learning and instinct applies more generally to phenotypic plasticity and phenotypic rigidity, but we will usually leave it to the reader to make this generalization, since it is not clear that the more general view is relevant to evolutionary computation.

## 4 Costs and Benefits of Phenotypic Plasticity and Rigidity

There is a common, seductive sentiment that learning is always good; that evolution always selects for more powerful, general-purpose learning engines. In fact, learning is not always advantageous. Learning is expensive, in many different ways. Evolution is constantly selecting the best balance between learning and instinct, and the best balance varies as behaviors evolve and the environment changes. There is growing evidence that the human brain has many more instinctive elements than we usually acknowledge (Barkow *et al.*, 1992; Pinker, 1994).

Another related myth is that the Baldwin effect is merely a kind of synergy effect; that the Baldwin effect is the synergy that results when learning (in individuals) is combined with evolution (in populations). This is only part of the truth. It is the first observation we listed in the preceding section. In fact, it is false unless it is carefully qualified. Learning can accelerate evolution under certain circumstances, but it can also slow evolution under other circumstances.

Baldwin (1896) proposed that learning (phenotypic plasticity) is advantageous when a new behavior is starting to evolve in a population. Learning smooths the fitness landscape, which facilitates evolution (observation 1). If it is possible for the behavior to be performed by an instinctive mechanism, it will usually be advantageous for such a mechanism to evolve, since instinctive mechanisms tend to be less expensive than learned mechanisms. However, when a new behavior is first evolving, an instinctive mechanism may require the population to make a large evolutionary leap, while a learned mechanism may be able to arise in smaller evolutionary increments. Learning may allow the behavior to eventually become common and robust in the population, which then gives evolution the time required to find an instinctive mechanism to replace the learned mechanism. In summary, at first learning is advantageous, but later it is not.

This picture still requires further qualification. The fundamental insight is that there are trade-offs between learning (plasticity) and instinct (rigidity). The optimal balance may vary over time, as the population and environment change. The precise course of this varying balance over time may not always follow the path that Baldwin (1896) described. The important lesson from Baldwin (1896) is not the precise course of the balance; it is that there are trade-offs. Table 1 is an attempt to list these trade-offs.

Table 1: Trade-offs in evolution between phenotypic rigidity and phenotypic plasticity.

| | dimension of trade-off | phenotypic rigidity (e.g., instinct) | phenotypic plasticity (e.g., learning) |
|---|---|---|---|
| 1 | time scale of environmental change | relatively static | relatively dynamic |
| 2 | variance, reliability | low variance, high reliability | high variance, low reliability |
| 3 | energy, CPU consumption | low energy, low CPU | high energy, high CPU |
| 4 | length of learning period | short learning period | long learning period |
| 5 | global versus local search | more global search | more local search |
| 6 | adaptability | brittle | adaptive |
| 7 | fitness landscape | rugged | smooth |
| 8 | reinforcement learning versus supervised learning | reinforcement learning | supervised learning |
| 9 | bias direction | strong bias: direction of bias crucial to success | weak bias: direction of bias not as important |
| 10 | global goals versus local goals | emphasis on global goals | emphasis on local goals |

**1. Time scale of environmental change:** Evolution and learning operate at different time scales (Unemi *et al.*,

1994; Anderson, 1995a). In a dynamic environment, evolution cannot adapt fast enough, so there is an advantage to learning (phenotypic plasticity). In a static environment, evolution can adapt, so there is no penalty for instinct (phenotypic rigidity), at least in this dimension (the time scale dimension).

**2. Variance and reliability:** Learning is based on experience and requires the right kind of experience. If an individual is unfortunate, the right experience will not be available. This factor makes learning more stochastic or probabilistic than instinct. If all the other factors are equal, evolution will eventually replace learning with instinct, simply because instinct is more reliable (Sober, 1994). On the positive side, learning can increase the variation in the population, which can facilitate evolution in some circumstances (such as a dynamic environment) (Anderson, 1995a).

**3. Energy consumption and CPU consumption:** Learning requires acquisition of data, which involves sensors and experiments. A living organism or a robot must expend energy in order to learn. In evolutionary computation, the CPU is a limited resource. Local search (learning) requires CPU time, which means that less CPU time is available for global search (evolution).

**4. Length of learning period:** An organism (or a robot) is vulnerable during the period before it has fully learned a certain behavior. For example, if the behavior is self-defense, the organism is easier to kill before it has mastered the behavior. If all the other factors are equal, evolution will select for shorter learning periods (Cecconi *et al.*, 1995; Anderson, 1995a).

**5. Global versus local search:** Evolution performs a kind of global search (Holland's Schema Theorem) while individual learning performs a kind of local search (in the phenotype space centered on a given genome). An organism with strong instincts is putting more emphasis on the global search, while an organism with weak instincts is putting more emphasis on the local search. The right trade-off depends on the fitness landscape and the current location of the population on that fitness landscape. Thus the right trade-off varies over the course of the evolution of a given behavior.

**6. Adaptability versus brittleness:** Learning is better able to adjust to variation in the environment. Instinct tends to be brittle.

**7. Fitness landscape:** Learning smooths the slope of the fitness landscape. If the slope is already smooth, learning may have little advantage over instinct (Hightower *et al.*, 1996). The relevant smoothness is the smoothness of the landscape around the current population location, which varies over the course of the evolution of a given behavior.

**8. Reinforcement versus supervised learning:** A genetic algorithm is a type of reinforcement learning algorithm. Therefore it is situated somewhere between unsupervised learning and supervised learning, in terms of its use of feedback from the environment. In a supervised learning task (e.g., learning to classify from examples), a standard supervised learner (e.g., backpropagation neural networks or decision tree induction) has an advantage over a reinforcement learner (e.g., a genetic algorithm), because the supervised learner uses more of the feedback from the environment. Suppose there are 10 classes and a learner mistakenly assigns an example to class 3 instead of class 8. A supervised learner can note that examples of this type should be assigned to class 8 in the future. A reinforcement learner can only note that examples of this type should not be assigned to class 3. A hybrid of a genetic algorithm and a supervised learning algorithm can have an advantage over a pure genetic algorithm when the environment provides detailed feedback (Nolfi *et al.*, 1994).

**9. Bias direction:** Bias is a familiar concept in machine learning: every inductive learner requires a bias in order to select one hypothesis from the infinite set of hypotheses that are consistent with a given set of observations (Haussler, 1988; Rendell, 1986; Utgoff, 1986). For example, a preference for simpler hypotheses is a form of bias. Bias has direction (correctness) and strength (Utgoff, 1986). There is a strong analogy between the learned-instinctive continuum and the strong-weak bias continuum. A strongly biased machine learning system is like an organism that emphasizes instinctive behaviors. A weakly biased machine learning system is like an organism that emphasizes learned behaviors. If the bias direction is correct (for example, if simpler hypotheses are more likely to be true than complex hypotheses), strong bias (instinct) is best, since a strong and correct bias accelerates learning (Utgoff, 1986). If the bias direction is incorrect (for example, there is no correlation between the complexity of hypotheses and the truth of hypotheses), weak bias (learning) is best, since a weak bias can be corrected with fewer data than a strong bias.

**10. Global goals versus local goals:** Evolution and learning have different goals. Evolution seeks to maximize fitness, but individuals have more immediate goals, such as to eat food that tastes good. Learning is used by individuals to help them achieve their immediate goals, which may not match with the goals of evolution (Nolfi *et al.*, 1994; Menczer & Belew, 1994; Turney, 1995). In biological evolution, no organism can have "maximize expected inclusive fitness" as a goal, because it is too difficult to determine whether a given action will contribute to this goal. Instead, biological organisms must substitute simpler goals, such as "seek sweet, fatty food". Whether these simpler goals will serve as reasonable substitutes for the goal of fitness is contingent on the environment. In evolutionary computation, evolution and learning can have identical goals. However, as the complexity of problems tackled by evolutionary computation increases, we may expect it to become more similar to biological evolution in this respect. For example, in machine learning, the problem of learning to classify accurately is simpler than the problem of learning to classify with low cost. One approach to classifying with low cost is to evolve a population of learners, where each individual has the goal of learning to classify accurately, but the fitness of the individuals is determined by the cost of classification (Turney, 1995).

The above list is not necessarily exhaustive and there may be some overlap in the items. The more we contemplate the Baldwin effect, the longer the list grows.

## 5 Lamarckism and the Baldwin Effect

In this section, we examine a cluster of myths and legends involving the relationship between Lamarckism and the Baldwin effect.

### 5.1 The Baldwin Effect is Purely Darwinian

Lamarck believed in the inheritance of acquired characteristics. In biology and in more complex evolutionary computation, there is a distinction between the genotype and the phenotype. Lamarckism requires an inverse mapping from phenotype and environment to genotype. This inverse mapping is biologically implausible. However, the Baldwin effect is purely Darwinian; not Lamarckian. The Baldwin effect does not involve any inverse mapping.

Suppose a short-necked animal learns to stretch its neck to reach nutritious leaves on a tall tree. Lamarck believed that the animal's offspring would inherit slightly longer necks than they would otherwise have had. This would require a mechanism for modifying the DNA of the parent, to alter its genes for neck length, based on its habit of stretching its neck.

The Baldwin effect has consequences that are similar to Lamarckian evolution. Over many generations, animals that stretch their necks may evolve longer necks. However, the mechanism is purely Darwinian. Parents who stretch their necks will pass on to their children *not* their longer necks, but rather their ability to stretch their necks.

Evolution will select for the ability to stretch. Over many generations, the population will evolve to consist largely of animals that are very good at stretching their necks. However, there can be advantages to being born with a longer neck. Given sufficient time, the population may eventually evolve longer necks. Their ability to stretch their necks is what grants them the time required to evolve longer necks. The point of this story is that the Baldwin effect is somewhat Lamarckian in its results, but it is not Lamarckian in its mechanisms.

### 5.2 Lamarckism is Computationally Intractable in General

It might be argued that, although Lamarckism is not biologically accurate, it is ideal for evolutionary computation. Living organisms do not modify their DNA, based on their experience, but we can simulate Lamarckian evolution in a computer. Perhaps Lamarckian evolution is superior to the Baldwin effect, when we are attempting to solve problems by evolutionary computation (Belew, 1990; Hightower *et al.,* 1996; Whitley *et al.,* 1994; Moscato, 1989, 1993; Moscato and Fontanari, 1990; Norman and Moscato, 1989; Moscato and Norman, 1992; Radcliffe and Surry, 1994; Paechter *et al*., 1995; Burke *et al.,* 1995).

Lamarckian evolution requires an inverse mapping from phenotype and environment to genotype. This inverse mapping may be computable in many simple applications of evolutionary computation. However, we believe that the computation will typically be intractable, for interesting, real-world problem solving. For example, we applied the Baldwin effect to the problem of learning to classify with low cost (Turney, 1995). The genotype was a string of bits, specifying a bias for a decision tree induction system. The phenotype was a decision tree. The environment was the data. The mapping from genotype (bias) and environment (data) to phenotype (decision tree) was easily computed, but there is no known algorithm for the inverse mapping from phenotype (decision tree) and environment (data) to genotype (bias). As our applications for evolutionary computation grow increasingly complex, Lamarckian evolution will become decreasingly feasible.

In some recent work with Lamarckian evolution, there is no distinction between the phenotype and the genotype (Whitley *et al*., 1994; Paechter *et al*., 1995; Burke *et al.,* 1995). This produces the misleading impression that the inverse mapping is trivial. It may well be trivial for many interesting and worthwhile problems, but we believe that it is generally intractable. With progress in evolutionary computing, we will eventually encounter the limits of the Lamarckian approach.

It has also been pointed out that Lamarckian evolution distorts the population so that the Schema Theorem no longer applies (Whitley *et al.*, 1994). The Baldwin effect alters the fitness landscape, but it does not modify the basic evolutionary mechanism (i.e., it is purely Darwinian). Therefore the Schema Theorem still applies to the Baldwin effect.

### 5.3 Memes are Not Necessarily Lamarckian

We would like to discuss one more myth concerning Lamarckian evolution and the Baldwin effect. Dawkins (1976) proposed that ideas evolve in culture in much the same sense as organisms evolve in biology, and he coined the term *meme* for the basic unit of cultural transmission, analogous to the gene in biological evolution. Dawkins (1982) and Gould (1991) have suggested that memes evolve by Lamarckian mechanisms. Several authors have used the term *memetic evolution* as essentially synonymous with Lamarckian evolution (Moscato, 1989, 1993; Moscato and Fontanari, 1990; Norman and Moscato, 1989; Moscato and Norman, 1992; Radcliffe and Surry, 1994; Paechter *et al.*, 1995; Burke *et al.,* 1995). Cziko (1995) argues to the contrary that meme evolution is purely Darwinian. We agree with Cziko (1995).

Let us examine some of the arguments for Lamarckian evolution of memes. To begin, we need to define the genotype-phenotype distinction for memes, since both Lamarckian evolution and the Baldwin effect require this distinction. The devices by which we express our ideas are analogous to genotypes; the ideas themselves are analogous to phenotypes. Since memes are defined as analogous to genes (Dawkins, 1976), we will use the term *meme* to refer to the devices by which we express our ideas. Examples of memes are spoken sentences, written sentences, live music, recorded music, theatre, and cinema. We will use the term *idea* to refer to the ideas themselves.

When a human brain receives a meme (i.e., is colonized by a meme), the meme slowly (over seconds or days) matures into an idea. Eventually the human may decide to communicate the idea to another person. Communication involves transmitting a meme. If memes use Lamarckian evolution, then there is a kind of *reverse engineering* in the host brain, by which the mature idea is transformed into a meme that captures the content of the mature idea. If memes use Darwinian evolution, then the meme that is transmitted is the result of mutation and crossover with other memes in the host brain. Introspection suggests that Lamarckian evolution is more accurate: memes appear to be a kind of *encoding* of mature ideas. However, introspection is notoriously unreliable. We will consider two other arguments for Lamarckian evolution of memes.

One argument for Lamarckian memes is based on comparing what people hear to what people say. The meme that is received by a brain is generally quite different from the meme that is later transmitted. We are not merely recording machines that can only play back what we have heard. However, this does not imply that memes are non-Darwinian. Suppose that a human brain colonized by memes is analogous to an island colonized by birds. The memes that shuttle back and forth from brain to brain are analogous to those birds that dare to leave their island of birth and fly to another island. If we see one type of bird fly to an island and then ten-thousand years later see a quite different type of bird leave the island, we would be wrong to infer that we had witnessed a case of Lamarckian evolution. It seems possible that memes in our brains may evolve as much in a few minutes or days as birds evolve in ten-thousand years.

A second argument for Lamarckian memes is based on creativity. Creative thought often seems to consist of combining ideas to make new ideas. This might appear to support a Lamarckian view. However, perhaps creative thought is a mating of memes, rather than a merging of ideas. Unlike biological organisms, memes do not seem to respect species boundaries; any two memes might mate with each other and produce fertile offspring. The evidence appears to be compatible with both Lamarck and Darwin.

We do not have proof that memes are Darwinian, nor that they are not Lamarckian. Our argument is that it is possible that memes are not Lamarckian. Therefore it is premature to use *memetic evolution* as a synonym for *Lamarckian evolution*.

### 5.4 Memes May Be Baldwinian

We believe that memes may evolve by exploiting the Baldwin effect. In support of this claim, we will argue that memes satisfy all of the necessary conditions for the manifestation of the Baldwin effect.

The Baldwin effect requires Darwinian evolution, which requires entities that reproduce, with heritable traits and some degree of variation, and selection, which typically arises from competition for limited resources. Dawkins (1976) has already argued persuasively that memes satisfy the requirements for Darwinian evolution. In addition, the Baldwin effect requires phenotypic plasticity. The population must display heritable variation in phenotypic plasticity, and plasticity must have costs and benefits, in terms of the mechanism of selection.

Ideas (the phenotypes of memes) clearly have varying degrees of plasticity. That is, some ideas are more flexible and adaptable than others. The environment to which ideas must adapt consists mainly of the other ideas that inhabit the host brain. The mechanism of selection is familiar, although poorly understood: we *choose* to entertain some ideas and ignore other ideas. In general, we choose plastic ideas; ideas that get along well with the other ideas we accept; ideas that fit into the ecology of the host brain. But plastic ideas also have costs. For an idea to adapt as a phenotype, during its lifetime in a particular host brain, the host brain must process the idea; the idea must consume brain time; the human host has to think hard. Therefore plasticity in ideas has both costs and benefits.

It appears that memes satisfy all of the requirements for manifestation of the Baldwin effect. We previously argued (Section 5.3) that memes are not necessarily Lamarckian. We believe that memes are more likely to be Baldwinian than Lamarckian. Our arguments do not prove this, but they at least show that Baldwinian memes are at least as plausible as Lamarckian memes.

Baldwinian and Lamarckian evolution are virtually indistinguishable in their effect. We believe that no "external" (e.g., linguistic) analysis of memes will be able to resolve the Baldwin/Lamarck meme dispute. It seems to us that only an "internal" (e.g., neurological) analysis can settle the arguments, just as Lamarck could only be properly rejected for biological evolution when the distinction between somatic and germ cells was discovered by the embryologist August Weismann.

## 6 Conclusion

We have argued that the Baldwin effect is widely misunderstood. There seem to be two general categories of misunderstandings. We discussed the first category in Section 4: many researchers have focused on the benefits of lifetime learning in an evolving population, but there are also costs. We listed some of the costs and benefits, but our list may be far from complete. The Baldwin effect has depths that we have not yet plumbed.

Section 5 examined the second category of myths and legends: those involving the relationship between Lamarckian evolution and the Baldwin effect. We argued three points: (1) The Baldwin effect is not Lamarckian. It is purely Darwinian. (2) There are reasons to believe that the Baldwin effect has more applications in evolutionary computation than Lamarckian evolution. Lamarckian evolution requires an inverse mapping from phenotype and environment to genotype. We believe that computing this mapping is intractable in general. (3) Contrary to popular opinion, it is not clear that memes use Lamarckian evolution. It is equally plausible that they use the Baldwin effect.

## Acknowledgments

Thanks to Russell Anderson, Darrell Whitley, and Joel Martin for inspiration and discussion. Thanks to Russell Anderson for helpful comments on an earlier version of this paper.

## References

Ackley, D., and Littman, M. (1991). Interactions between learning and evolution. In *Proceedings of the Second Conference on Artificial Life*, C. Langton, C. Taylor, D. Farmer, and S. Rasmussen, editors. California: Addison-Wesley.

Anderson, R.W. (1995a). Learning and evolution: A quantitative genetics approach. *Journal of Theoretical Biology*, 175, 89-101.

Anderson, R.W. (1995b). Genetic mechanisms underlying the Baldwin effect are evident in natural antibodies. In *Evolutionary Programming IV: The Edited Proceedings of the Fourth Annual Conference on Evolutionary Programming*, edited by J.R. McDonnell, R.G. Reynolds, and D.B. Fogel, pp. 547-563. Cambridge, MA: MIT Press.

Balakrishnan, K., and Honavar, V. (1995). Evolutionary design of neural architectures: A preliminary taxonomy and guide to literature. Artificial Intelligence Research Group, Department of Computer Science, Iowa State University, Technical Report CS TR #95-01.

Baldwin, J.M. (1896). A new factor in evolution. *American Naturalist*, 30, 441-451.

Barkow, J.H., Cosmides, L., and Tooby, J. (1992). Editors, *The Adapted Mind: Evolutionary Psychology and the Generation of Culture,* New York: Oxford University Press.

Behera, N., and Nanjundiah, V. (1995). An investigation into the role of phenotypic plasticity in evolution. *Journal of Theoretical Biology*, 172, 225-234.

Belew, R.K. (1989). When both individuals and populations search: Adding simple learning to the Genetic Algorithm. In *Proceedings of the Third International Conference on Genetic Algorithms*, 34-41, Washington D.C.

Belew, R.K. (1990). Evolution, learning and culture: computational metaphors for adaptive search. *Complex Systems,* 4, 11-49.

Belew, R.K., McInerney, J., and Schraudolph, N.N. (1991). Evolving networks: Using the Genetic Algorithm with connectionist learning. In *Proceedings of the Second Artificial Life Conference,* 511-547, Addison-Wesley.

Belew, R.K. and Mitchell, M., (1996). Editors, *Adaptive Individuals in Evolving Populations: Models and Algorithms,* Massachusetts: Addison-Wesley.

Burke, E.K., Newall, J.P., and Weare, R.F. (1995). A memetic algorithm for university exam timetabling. *Proceedings of the First International Conference on the Practice and Theory of Automated Timetabling (ICPTAT-95),* 496-503. Napier University, Edinburgh.

Cecconi, F., Menczer, F., and Belew, R.K. (1995). Maturation and the evolution of imitative learning in artificial organisms. Technical Report

CSE 506, University of California, San Diego, 1995; to appear in *Adaptive Behavior*, 4, January 1996.

Cziko, G. (1995). *Without Miracles: Universal Selection Theory and the Second Darwinian Revolution*. Massachusetts: MIT Press.

Dawkins, R. (1976). *The Selfish Gene*. Oxford: Oxford University Press.

Dawkins, R. (1982). *The Extended Phenotype: The Gene as the Unit of Natural Selection.* San Francisco: Freeman.

French, R., and Messinger, A. (1994). Genes, phenes and the Baldwin effect. In Rodney Brooks and Patricia Maes (editors), *Artificial Life IV*. Cambridge MA: MIT Press.

Gottlieb, G. (1992). Evolution: The modern synthesis and its failure to incorporate individual development into evolutionary theory. Chapter 11 of *Individual Development and Evolution: The Genesis of Novel Behavior*. New York: Oxford University Press.

Gould, (1991). *Bully for Brontosaurus: Reflections in Natural History*. New York: Norton.

Hart, W.E. (1994). *Adaptive Global Optimization with Local Search*. Ph.D. Thesis, Department of Computer Science and Engineering, University of California, San Diego.

Hart, W.E., and Belew, R.K. (1996). Optimization with genetic algorithm hybrids that use local search. In R.K. Belew and M. Mitchell, (editors), *Adaptive Individuals in Evolving Populations: Models and Algorithms,* Addison-Wesley.

Hart, W.E., Kammeyer, T.E., and Belew, R.K. (1995). The role of development in genetic algorithms. In L.D. Whitley and M.D. Vose, (editors), *Foundations of Genetic Algorithms 3,* California: Morgan Kaufmann.

Harvey, I. (1993). The puzzle of the persistent question marks: A case study of genetic drift. In S. Forrest (editor) *Proceedings of the Fifth International Conference on Genetic Algorithms, ICGA-93*, California: Morgan Kaufmann.

Haussler, D. (1988). Quantifying inductive bias: AI learning algorithms and Valiant's learning framework. *Artificial Intelligence,* 36, 177-221.

Hightower, R., Forrest, S., and Perelson, A. (1996) The Baldwin effect in the immune system: learning by somatic hypermutation. In R.K. Belew and M. Mitchell, (editors), *Adaptive Individuals in Evolving Populations: Models and Algorithms,* Addison-Wesley.

Hinton, G.E., and Nowlan, S.J. (1987). How learning can guide evolution. *Complex Systems,* 1, 495-502.

Maynard Smith, J. (1987). When learning guides evolution. *Nature,* 329, 761-762.

Menczer, F., and Belew, R.K. (1994). Evolving sensors in environments of controlled complexity. In R. Brooks and P. Maes, (editors), *Artificial Life IV,* MIT Press.

Morgan, C.L. (1896). On modification and variation. *Science,* 4, 733-740.

Moscato, P. (1989). On evolution, search, optimization, genetic algorithms and martial arts: Towards memetic algorithms. Technical Report 826, California Institute of Technology, Pasadena, California.

Moscato, P. (1993). An introduction to population approaches for optimization and hierarchical objective functions: The role of tabu search. *Annals of Operations Research,* 41, 85-121.

Moscato, P., and Fontanari, J.F. (1990). Stochastic versus deterministic update in simulated annealing. *Physics Letters A*, 146, 204-208.

Moscato, P. and Norman, M.G. (1992). A memetic approach for the travelling salesman problem. In *Proceedings of the International Conference on Parallel Computing and Transputer Applications*, 177-186. Amsterdam: IOS Press.

Nolfi, S., Elman, J.L., and Parisi, D. (1994). Learning and evolution in neural networks, *Adaptive Behavior,* 3, 5-28.

Norman, M.G., and Moscato, P. (1989). A competitive-cooperative approach to complex combinatorial search. In *Selected Work for the Proceedings of the 20th Joint Conference on Informatics and Operations Research*, 3.15-3.29, Buenos Aires, Argentina.

Osborn, H.F. (1896). Ontogenic and phylogenic variation. *Science,* 4, 786-789.

Paechter, B., Cumming, A., Norman, M.G., and Luchian, H. (1995). Extensions to a memetic timetabling system. *Proceedings of the First International Conference on the Practice and Theory of Automated Timetabling (ICPTAT-95),* 455-467. Napier University, Edinburgh.

Pinker, S. (1994). *The Language Instinct: How the Mind Creates Language*, New York: William Morrow and Co.

Radcliffe, N.J., and Surry, P.D., (1994). Formal memetic algorithms. *Proceedings of the AISB Workshop on Evolutionary Computing.* Berlin: Springer-Verlag.

Rendell, L. (1986). A general framework for induction and a study of selective induction. *Machine Learning*, 1, 177-226.

Scheiner, S. (1993). Genetics and evolution of phenotypic plasticity. *Annual Review of Ecology and Systematics,* 24, 35-68.

Simpson, G.C. (1953). The Baldwin effect. *Evolution,* 7, 110-117.

Sober, E. (1994) The adaptive advantage of learning and a priori prejudice. In *From a Biological Point of View: Essays in Evolutionary Philosophy*, a collection of essays by E. Sober, 50-70, Cambridge University Press.

Turney, P.D. (1995). Cost-sensitive classification: Empirical evaluation of a hybrid genetic decision tree induction algorithm. *Journal for AI Research*, 2, 369-409.

Unemi, T., Nagayoshi, M., Hirayama, N., Nade, T., Yano, K., and Masujima, Y. (1994). Evolutionary differentiation of learning abilities - a case study on optimizing parameter values in Q-learning by a genetic algorithm, In Rodney Brooks and Patricia Maes (editors), *Artificial Life IV.* Cambridge MA: MIT Press.

Utgoff, P. (1986). Shift of bias for inductive concept learning. In *Machine Learning: An Artificial Intelligence Approach, Volume II*. Edited by R.S. Michalski, J.G. Carbonell, and T.M. Mitchell. California: Morgan Kaufmann.

Waddington, C.H. (1942). Canalization of development and the inheritance of acquired characters. *Nature,* 150, 563-565.

Wcislo, W.T. (1989). Behavioral environments and evolutionary change. *Annual Review of Ecology and Systematics,* 20, 137-169.

West-Eberhard, M. (1989). Phenotypic plasticity and the origins of diversity. *Annual Review of Ecology and Systematics,* 20, 249-278.

Whitley, D., and Gruau, F. (1993). Adding learning to the cellular development of neural networks: Evolution and the Baldwin effect. *Evolutionary Computation*, 1, 213-233.

Whitley, D., Gordon, S., and Mathias, K. (1994). Lamarckian evolution, the Baldwin effect and function optimization. *Parallel Problem Solving from Nature - PPSN III.* In Y. Davidor, H.P. Schwefel, and R. Manner, editors, pp. 6-15. Berlin: Springer-Verlag.